\documentclass[conference]{IEEEtran}

\IEEEoverridecommandlockouts

\usepackage{algorithmic}
\usepackage{amsfonts}
\usepackage{amsmath}
\usepackage{amssymb}
\usepackage{cite}
\usepackage{comment}
\usepackage[T1]{fontenc}
\usepackage{graphicx}
\usepackage[super]{nth}
\usepackage{textcomp}
\usepackage{xcolor}
\usepackage{verbatim}
\usepackage{stfloats}
\usepackage{subcaption}
\usepackage[font=footnotesize]{caption}
\usepackage[font=footnotesize]{subcaption}
\usepackage{hyperref}

\begin{document}

\title{\LARGE \bf Shared-Control Robotic Manipulation in Virtual Reality}

\author{Shiyu Xu, Scott Moore and Akansel Cosgun \\
Monash University, Australia}

\maketitle

\begin{abstract}
In this paper, we present the implementation details of a Virtual Reality (VR)-based teleoperation interface for moving a robotic manipulator. We propose an iterative human-in-the-loop design where the user sets the next task-space waypoint for the robot's end effector and executes the action on the physical robot before setting the next waypoints. Information from the robot's surroundings is provided to the user in two forms: as a point cloud in 3D space and a video stream projected on a virtual wall. The feasibility of the selected end effector pose is communicated to the user by the color of the virtual end effector. The interface is demonstrated to successfully work for a pick and place scenario, however, our trials showed that the fluency of the interaction and the autonomy level of the system can be increased.
\end{abstract}


\section{Introduction}

The ultimate goal for robotics is to have fully autonomous agents that operate alongside humans. While there has been tremendous advances in robotics technology in recent years, the applications where autonomous robots can be deployed are still quite limited. In these cases, remotely controlled, or \textit{teleoperated}, human-in-the-loop robotics solutions may be viable. Teleoperated robots can perform tasks which are physically infeasible or inconvenient for humans, such as search-and-rescue operations or nuclear decommissioning. Capable of being quickly deployed and operated in hazardous environments, teleoperated robots have numerous practical applications that can make work safer and more efficient. In recent works in the field, such robot systems are often controlled by joysticks or other commercially available input devices designed for video games, with sensor data visualized on monitor displays \cite{tanaka2018development, klamt2018supervised, megalingam2018ros}. Recently competitions organized by DARPA exclusively featured robots that are remotely controlled \cite{hudson2021heterogeneous, atkeson2018happened}. While all teams employed some level of autonomy, there was no consensus on what the right autonomy level was \cite{atkeson2018happened}. The traditional monitor, keyboard and mouse interface was standard in these competitions. While having a monitor interface has its advantages, Augmented Reality (AR) and Virtual Reality (VR) offer a promising alternative as immersive interfaces that allows operation in three-dimensional space. 

The use of AR/VR to assist robot navigation has been embraced by researchers in an attempt to improve the user experience of robot teleoperation. Improvements in the performance and affordability of AR and VR devices have made them increasingly more suitable for the teleoperation of robots. Previous studies have shown that these interfaces can provide a significant advantage in both the spatial awareness and immersion of a user when compared with traditional techniques \cite{stotko2019vr, whitney2020comparing}. AR and VR also open the door to unique experiences which were not possible with previous technologies \cite{lemasurier2021semi, hetrick2020comparing}. While AR is more suitable for interfacing with co-located robots \cite{hoang2022arviz, gu2021seeing,waymouth2021demonstrating, newbury2021visualizing,gu2022ar}, VR is arguably better suited for teleoperation.

\begin{figure}[t!]
    \includegraphics[width=\columnwidth]{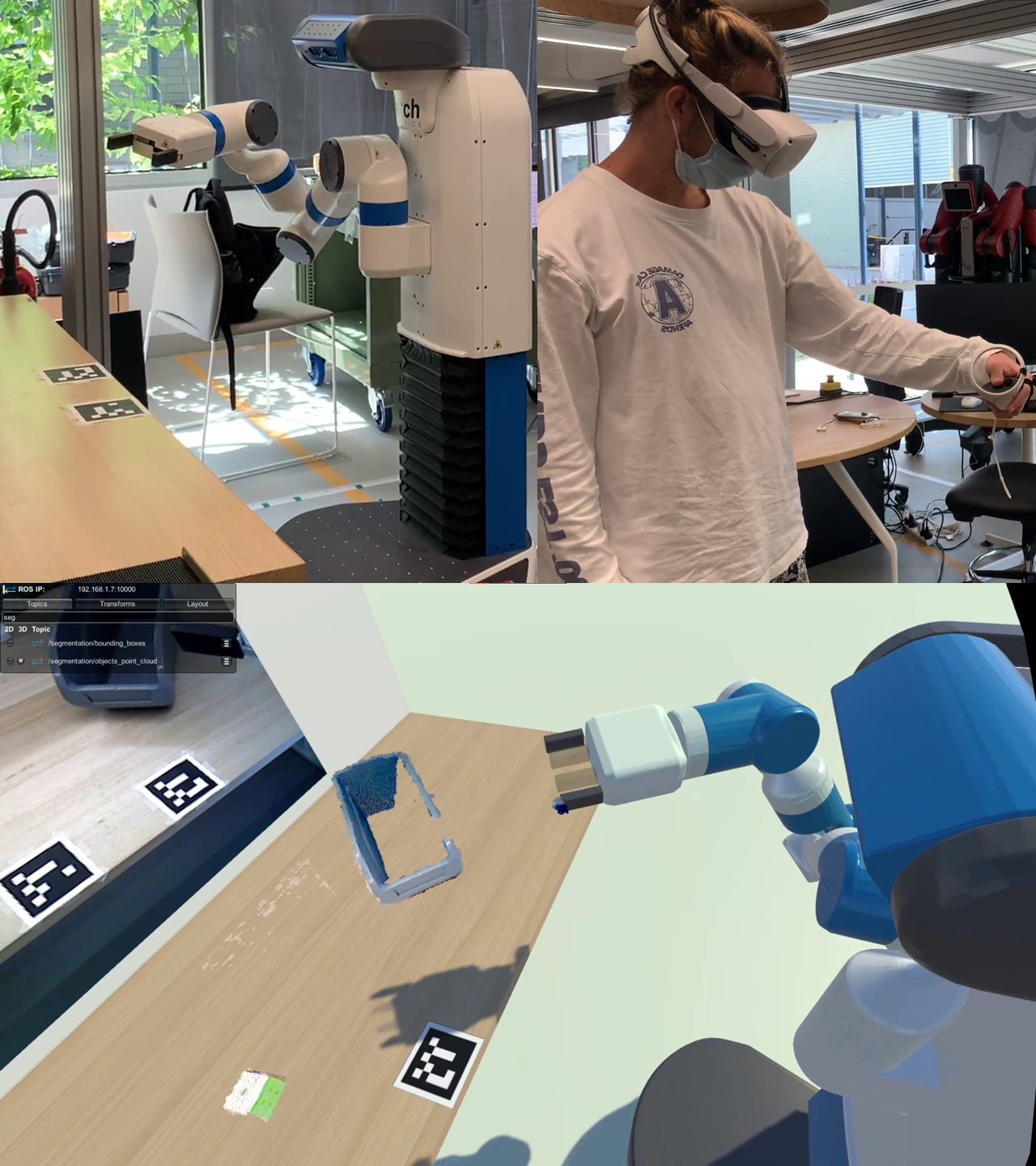}
    \caption{Our shared-control VR interface in action. A robot manipulator (top left) is controlled by a remote operator using a Virtual Reality headset (top right) and hand-held controllers. The operator sees a virtual representation of the environment including sensor information (bottom).}
    \label{figure:intro}
\end{figure}



Various different control methods have been used for teleoperated robotic manipulation. Waypoint-based motion planning allows a user to set sequential positions for the end effector, which the robot will attempt to adhere to when planning the movement of its arm. This approach is frequently used due to its simplicity and familiarity through traditional interfaces provided by the likes of MoveIt's \cite{coleman2014moveit} plugin in the Rviz visualization package. Other attempted methods in VR include trajectory control, where the end effector's movement mimics a recorded path of the user's hand in motion. Hetrick et al. \cite{hetrick2020comparing} compared the efficacy of such positional (waypoint-like) control against trajectory control, their user studies concluded that although no consistent differences were found in the users' reported experiences, the waypoint based control scheme achieved better completion time and accuracy in manipulation tasks. LeMasurier et al. \cite{lemasurier2021semi} designed an interface to create, interact and visualize functional waypoints to control the arm and gripper of a Fetch mobile manipulator. Wonsick et al. implemented a similar approach of waypoint based manipulation for a humanoid Valkyrie robot \cite{wonsick2021human}, and for a Toyota Human Support Robot in a separate study \cite{wonsick2021telemanipulation}.  

In regards to the visual presentation adopted by other research, it is common to allow user to freely move around the robot in an extrinsic manner and inspect the robot from any desired angle or position \cite{lemasurier2021semi, wonsick2021human, wonsick2021telemanipulation}. Sensory inputs from the robot is often shared with the user in the form of images \cite{macias2019measuring}, point clouds in a virtual environment \cite{wonsick2021human} or a combination of both \cite{lemasurier2021semi}. Wei et al. studied a merged visualization using point clouds projected on to an image, as well as a three-dimensional point cloud representation \cite{wei2021multi}. These approaches were compared to a picture-in-picture image display, which displays two two-dimensional images from different angles, one of them in a smaller frame on top of the other. Results from the subsequent user study suggest higher efficiency from the methods involving point clouds. The traditional point cloud visualization was compared by Wonsick et al. to a recreated virtual environment comprising of object models classified and localised by a deep learning model \cite{wonsick2021telemanipulation}. It was concluded that the recreated model visualizations was beneficial to the subjective user experience. 

In this paper, we develop a VR-based teleoperation interface for robot arm manipulation, using common practises for control and visualization. The operator wears a VR headset (Meta Quest 2) to move the manipulator of a Fetch robot platform. In our design, information coming from the RGB-D camera output located at the Fetch robot's head is displayed in two forms: the RGB video stream is projected to a virtual wall and the point cloud is displayed at the matching positions. We propose an iterative human-in-the-loop design where the user sets the next waypoint for the robot's end effector. The user controls the robot by selecting a 6D end-effector pose in the task space, as well as the state of the gripper (open or closed). The feasibility of the desired end effector pose is communicated to the user by the color the virtual end effector. The back-end utilises MoveIt for motion planning and default joint position control provided by the robot's low-level drivers.

The contributions of this work is two-fold:
\begin{itemize}
    \item Integration of Meta Quest 2 VR headset and Fetch robot for a successful demonstration of remote manipulation\footnote{\url{https://youtu.be/GDMIsA0xX5k}}.
    \item We release the source code of the project\footnote{\url{https://github.com/scottwillmoore/fetch\_vr}} so that other robotics researchers can replicate and build on top of our system.
\end{itemize}

\section{System and environment}

\begin{figure*}[t!]
\centering
\includegraphics[width=0.95\textwidth]{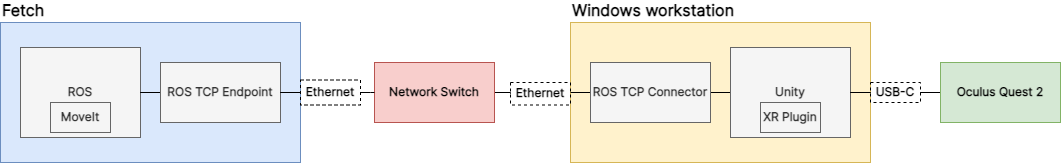}
\caption{Modules of the system. Software components are in grey, physical devices are colored and connection interfaces are enclosed in dotted lines.}
\label{figure:system}
\end{figure*}

\subsection{System}

The robot used was the Fetch mobile manipulator robot by Fetch Robotics. It features a 7-DOF arm and a head-mounted RGB-D camera capable of capturing 2D color images and 3D point clouds. Though the robot is capable of mobile navigation, in the implementation of this work its base is kept static. An on-board computer is capable of running Robot Operating System (ROS) \cite{quigley2009ros} on the installed Ubuntu 18.04 operation system, and also able to communicate with other devices via wireless and wired networks. The robot runs its own low-level software in the background to constantly publish sensor information onto ROS topics and execute incoming movement commands for its arm. Higher level tasks such as processing point cloud data are implemented in our own developed package which also run on the Fetch's on-board processor.

The VR device used was the commercially available Meta Quest 2, which includes a headset and a pair of hand-held wireless Meta Touch controllers. Both controllers can be tracked with 6-DOF, relative to the headset. In addition, Meta does provide support for hand and finger tracking, however these features were not used in our application. On each controller, there are four buttons and a joystick as shown in Fig.~\ref{figure:controller}. Through a USB-C connection, the headset was connected to and controlled by a workstation PC, instead of executing programs on the Quest 2's on-board processor. This was done to improve the development experience, and to reduce computational latency to avoid nausea induced by slow updates to the headset display.

\begin{figure}
\includegraphics[width=\columnwidth]{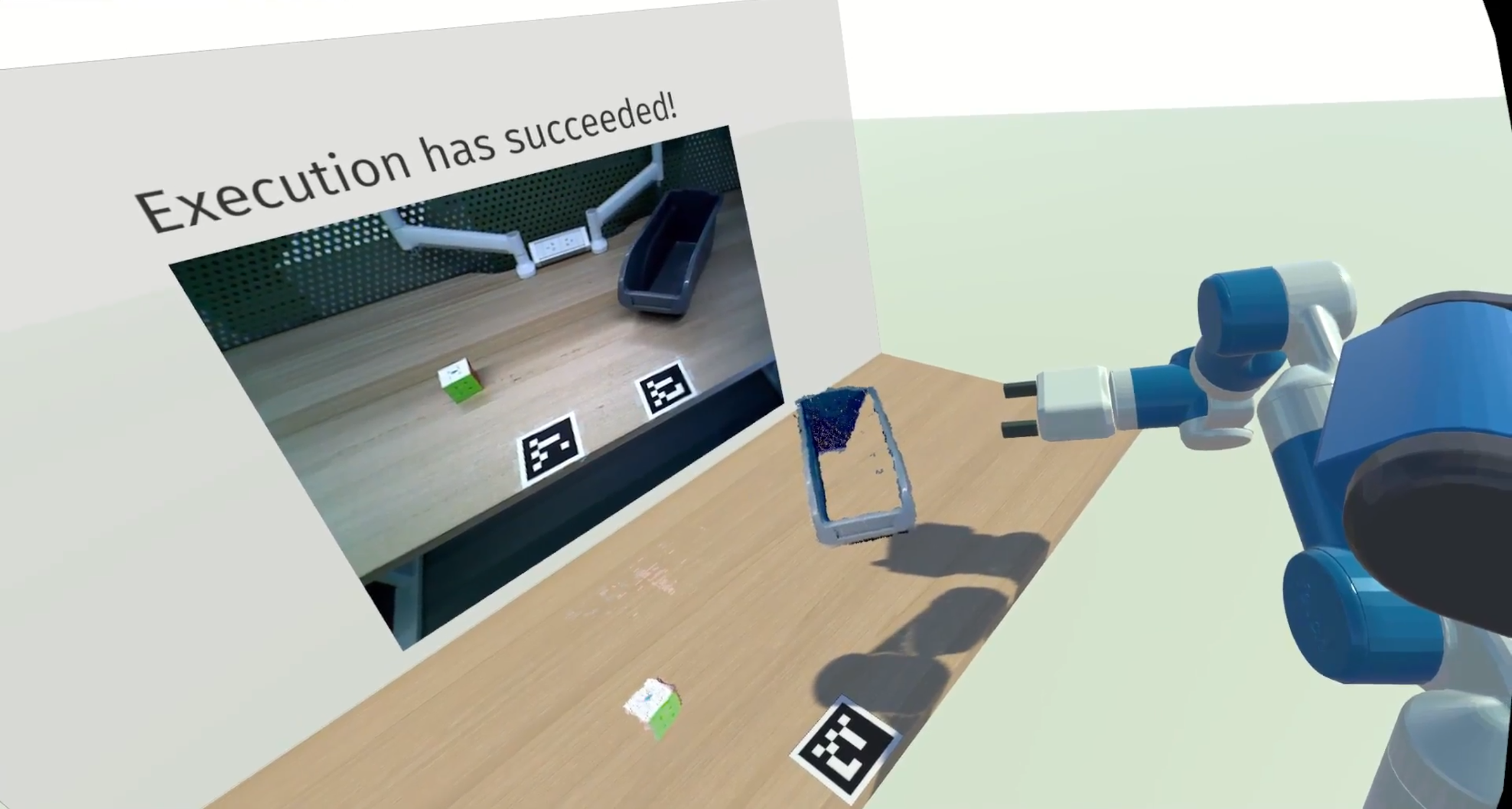}
\caption{The virtual workspace in the Unity application, with a model of the Fetch, synchronised with the physical robot, positioned in front of a desk workspace.}
\label{figure:application}
\end{figure}

Unity was used to build the virtual environment displayed to the user, shown in Figure \ref{figure:application}. While Unity is cross-platform, the Meta Quest 2 drivers are only available for Windows. Therefore, the VR component was developed and run on a separate workstation computer with Windows installed. This is the workstation connected to the VR headset. To relay communication between the Unity application (Windows) to the robot (Ubuntu), the packages provided by the Unity robotics \cite{unity2022} team were used. This utilises the Unity plugin, ROS TCP connector, which communicates with the ROS package, ROS TCP endpoint. Both wrap ROS messages into lightweight binary-encoded packets, which is more efficient \cite{tcp2022} than other methods, such as the JSON/BSON-encoded ROS Bridge \cite{rosbridge2022}.

We decided to connect all devices through Ethernet to a switch, to increase the available bandwidth of the system. It was found that WiFi was not able to provide the necessary throughput required to transfer point cloud data in real-time. A comprehensive diagram outlining the connected system can be seen in Figure \ref{figure:system}. The wired network was only possible due to the omission of movement of the robot's mobile base.

\subsection{Environment}

\begin{figure}
\includegraphics[width=\columnwidth]{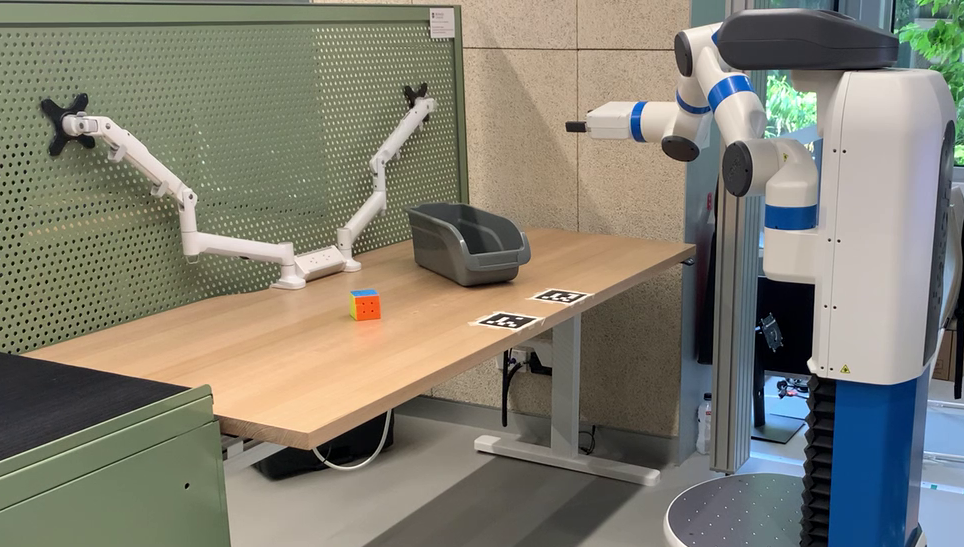}
\caption{The physical robot workspace.}
\label{figure:environment}
\end{figure}

The physical space allocated for the task was a desk whereby the robot was placed in front in roughly the same position for each test, as seen in Figure \ref{figure:environment}. The shape of the desk was modelled as a box, and added to the MoveIt planning scene as a collision object. The other objects that were placed on top of the table to interact with were not modelled, which therefore left the user responsible for avoidance.

\section{Visualization and motion planning}

\subsection{Visualization} \label{visualization}

The visualized components in the virtual environment consist of: a desk object modelled in Unity to match real life appearance, a point cloud of items on the desk, and a head-up display (HUD).

The point cloud detected by the Fetch's RGB-D camera is colorized and published by its internal software. As the network bandwidth is limited, the number of points was reduced as much as possible to maximise the refresh rate of the point cloud. As objects that do not move could be easily visualized by virtual models, only objects which were expected to change position or orientation were included. These objects, such as items on the desk, would be moved as a part of the pick-and-place task of the demonstration and thus need to be constantly updated. Static objects such as the desk, walls and the floor, were detected as large planes in the point cloud and removed. This was achieved with the 3D data processing library Open3D in Python, which provides a plane segmentation function on point cloud data using the random sample consensus (RANSAC) algorithm. Though simple, this implementation is prone to removing objects on the table which has relatively large planar surfaces, such as the target box where picked items are dropped into. Distant portions outside of the robot's interactive area were also removed.

\begin{figure}
    \includegraphics[width=\columnwidth]{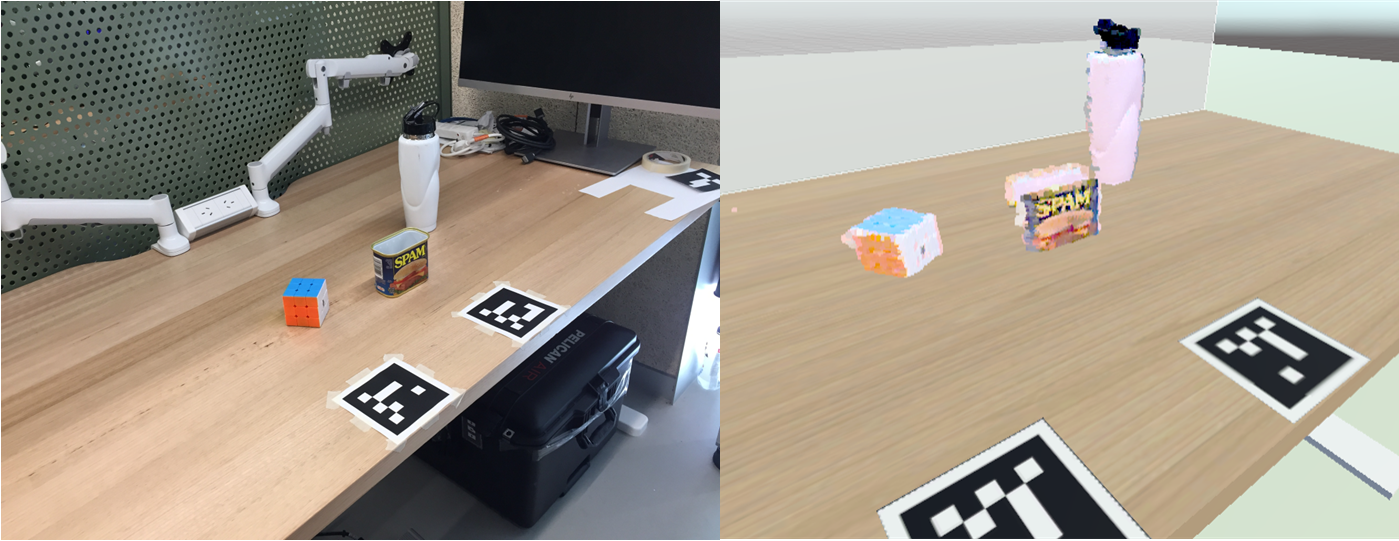}
    \caption{Point clouds of objects on the model desk, alongside a photo of the physical environment. The markers were rendered in the virtual world for the purpose of debugging their positions.}
    \label{figure:pointcloud}
\end{figure}

To recreate the physical environment in virtual form to aid the sense of immersion, a desk model was built from Unity cube objects with texture applied. It is localised by two ArUco markers \cite{garrido2014automatic} (only one of which was required and used), which can be detected with their position and orientation relative to the robot's camera. Thus by offsetting the coordinates of the marker, it is possible to obtain the location of the desk in the physical environment as relative to the robot, and replicate this in the virtual environment. Processed point clouds of objects and the model desk are shown in Figure \ref{figure:pointcloud}.

\begin{figure}
    \includegraphics[width=\columnwidth]{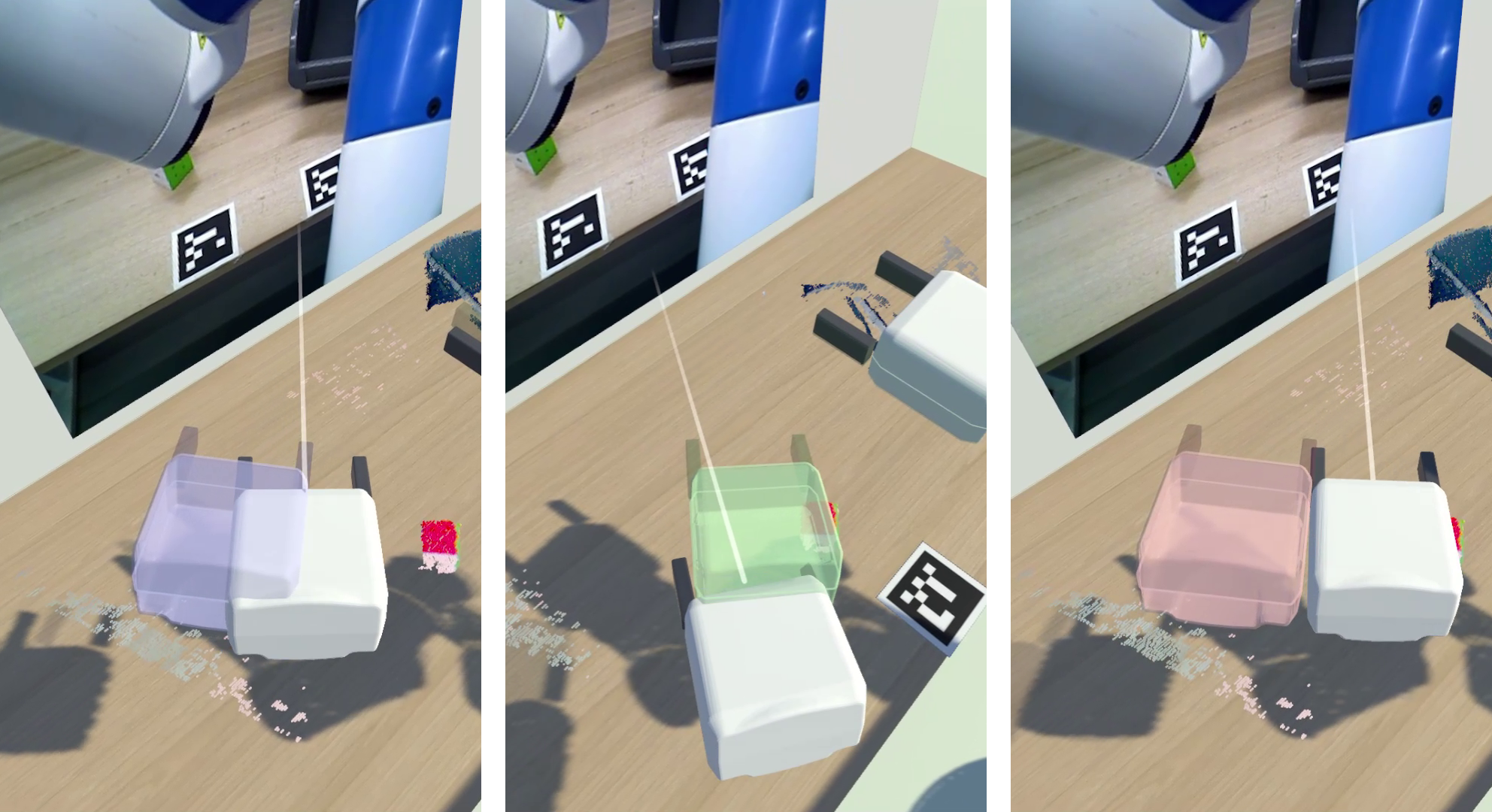}
    \caption{Plan feedback indicators represented as colored gripper end effectors. The position and orientation of the indicator represents that of the planned goal. Respectively, their colors represent a pending planning request (blue), a successful plan (green) and a failed plan (red).}
    \label{figure:feedback}
\end{figure}

The HUD, seen in Figure \ref{figure:application} which sits on a static vertical wall behind the virtual desk, displays the live 2D camera footage from the Fetch. This provides an alternative viewing option with higher clarity and consistency in tandem with the more spatially informative point cloud display. This view is especially necessary when the arm of the Fetch is obstructing its camera, a common issue when simultaneously operating the robot's arm and camera. When this happens the point cloud of obstructed objects are no longer visible. Why this occurs may not be immediately obvious if only the point cloud is presented, especially to inexperienced users. Additionally, a text-based feedback message of the motion plan is also provided above the camera display. However in trials it was found that this option was inconvenient as users would have to constantly move their sight between the planned goal and the HUD. During the UI design process, the text-based information display was later replaced with an indicator in the form of a colored, semi-transparent gripper end effector, with examples shown in Figure \ref{figure:feedback}.


An experimental effort was also made to improve the comprehensibility of the point cloud by clustering points belonging to the same object and enclosing them with rectangular 3D bounding boxes. A major challenge faced when attempting this was the limited completeness of the objects in point cloud form. As the robot was fixed in place and could only observe the scene from one angle, the point cloud only represented the objects from that angle. Thus the bounding boxes were shaped around the partial point clouds rather than that of the actual objects, offering little assistance in identifying the objects. The program behind this was also computationally intensive and drastically reduced the update rate of the point cloud. This was included as an optional feature in the final build. 

\subsection{Motion planning}

To plan and execute joint trajectories, the MoveIt motion planning framework was utilised. It was decided that the majority of the logic would be implemented on the Unity-side, as this would make it easier to access and display feedback to the user in the virtual environment.

MoveIt provides ROS actions and topics which could be interacted with from Unity with the assistance of the Unity robotics plugins. There was no implementation of a ROS action client for Unity, therefore a minimal, compliant action client was developed so that actions could be executed across the boundary.

The Unity plugin, ROS TCP Connector, which facilitated communication between Unity and ROS, also allows specification of coordinate frame data as following the ROS standard FLU definition (the x,y,z axes point forward, left, up respectively) or Unity's RUF definition (right, up, forward). The plugin also allows easy conversions of coordinates between the two frames. Therefore, ROS and Unity can share coordinate frames, whereby the origin of the virtual world is placed at the base of the robot. When the user attempts to create a plan, the request is sent to MoveIt with the \textit{MoveGroup} action, and Unity then awaits the response. To execute a plan, the \textit{ExecuteTrajectory} action is sent to MoveIt.

\subsection{User control}

\begin{figure}
\centering
    \begin{subfigure}[c]{0.48\columnwidth}
        \includegraphics[width=\columnwidth]{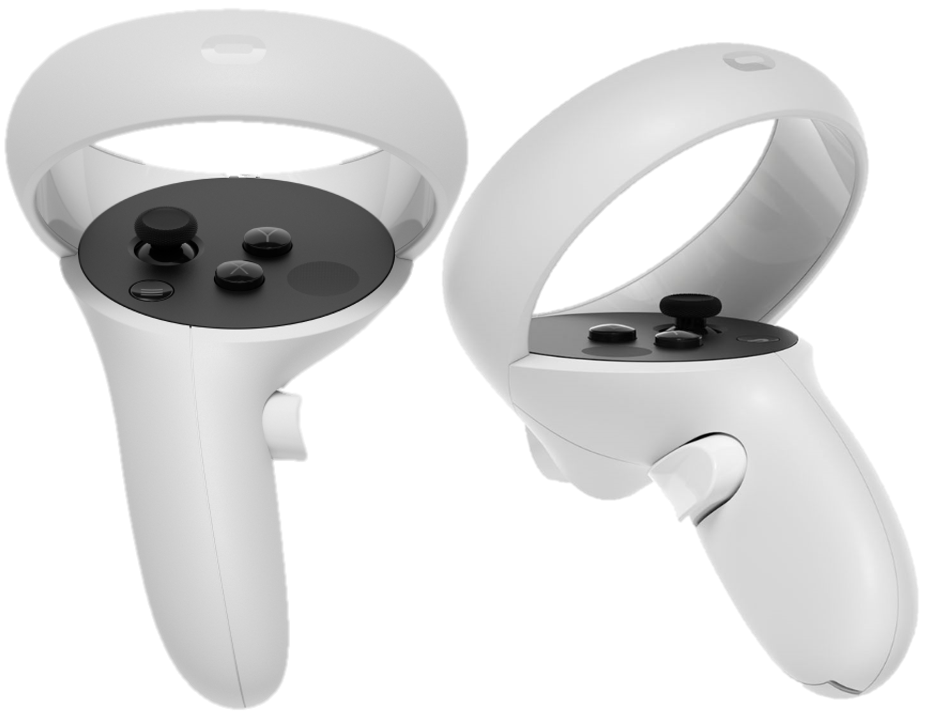}
        \caption{The Touch controllers}
        \label{figure:controller}
    \end{subfigure}
    \begin{subfigure}[c]{0.48\columnwidth}
        \includegraphics[width=\columnwidth]{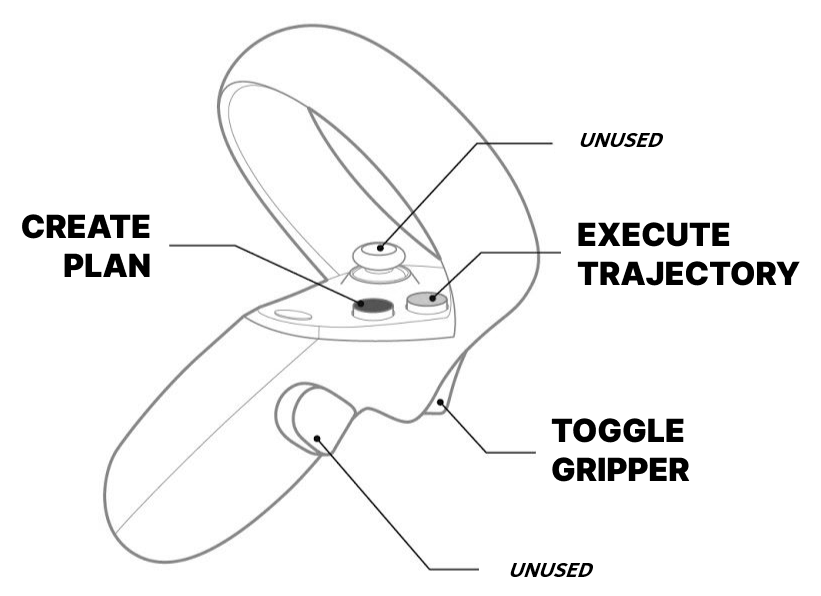}
        \caption{Controller button mapping}
        \label{figure:buttons}
    \end{subfigure}
    \caption{The controller layout. The left and right controller button mapping mirror each other.}
\end{figure}

The user provides commands to the robot through buttons on the Meta Quest 2 Touch controllers, shown in Figure \ref{figure:controller}. While the Fetch only has a single arm, both controllers can be used to control the robot. The user is free to use their dominant hand, or both hands for better reach and orientation. The button layout can be seen in Figure \ref{figure:buttons}.

The control method was designed for simplicity and immersion. The user is placed in the virtual world where they can see both the robot and the table. The robot state, which consists of the 7 joint angles and the gripper width, is synchronised with the real Fetch robot. They are free to walk around the environment to view the scene from a desired angle, as long as the user stays within the boundaries of the Meta Quest 2's play area called the \textit{Guardian}, which is set by the user before the experiment starts. Each controller is rendered as the end effector of the robot, instead of the rendering of the standard Oculus Touch controller model.

When the user moves their hand holding the controller, the virtual end effector follows the position and orientation of the controller. This does not trigger any movement of the robot. On the press of the \textit{Create Plan} button, the application attempts to find a plan from the robot's current state to the position and orientation of the controller where the button was pressed. A semi-transparent copy of the end effector is placed and it is colored based on the state of the request as seen in Figure \ref{figure:feedback}. Blue indicates a pending request, green indicates a successful plan has been made and red indicates that no plan could be found.

\begin{figure}
    \centering
    \begin{subfigure}[c]{0.255\columnwidth}
        \centering
        \includegraphics[width=\columnwidth]{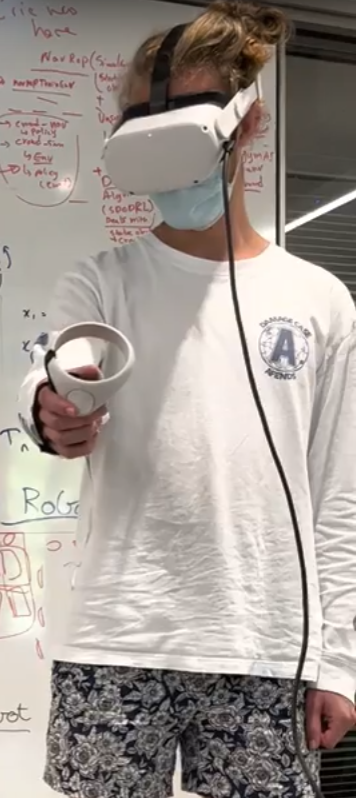}
        \caption{Operator}
        \label{figure:plan_a}
    \end{subfigure}
    \begin{subfigure}[c]{0.7\columnwidth}
        \centering
        \includegraphics[trim=0 0 0 0.05cm, clip,width=\columnwidth]{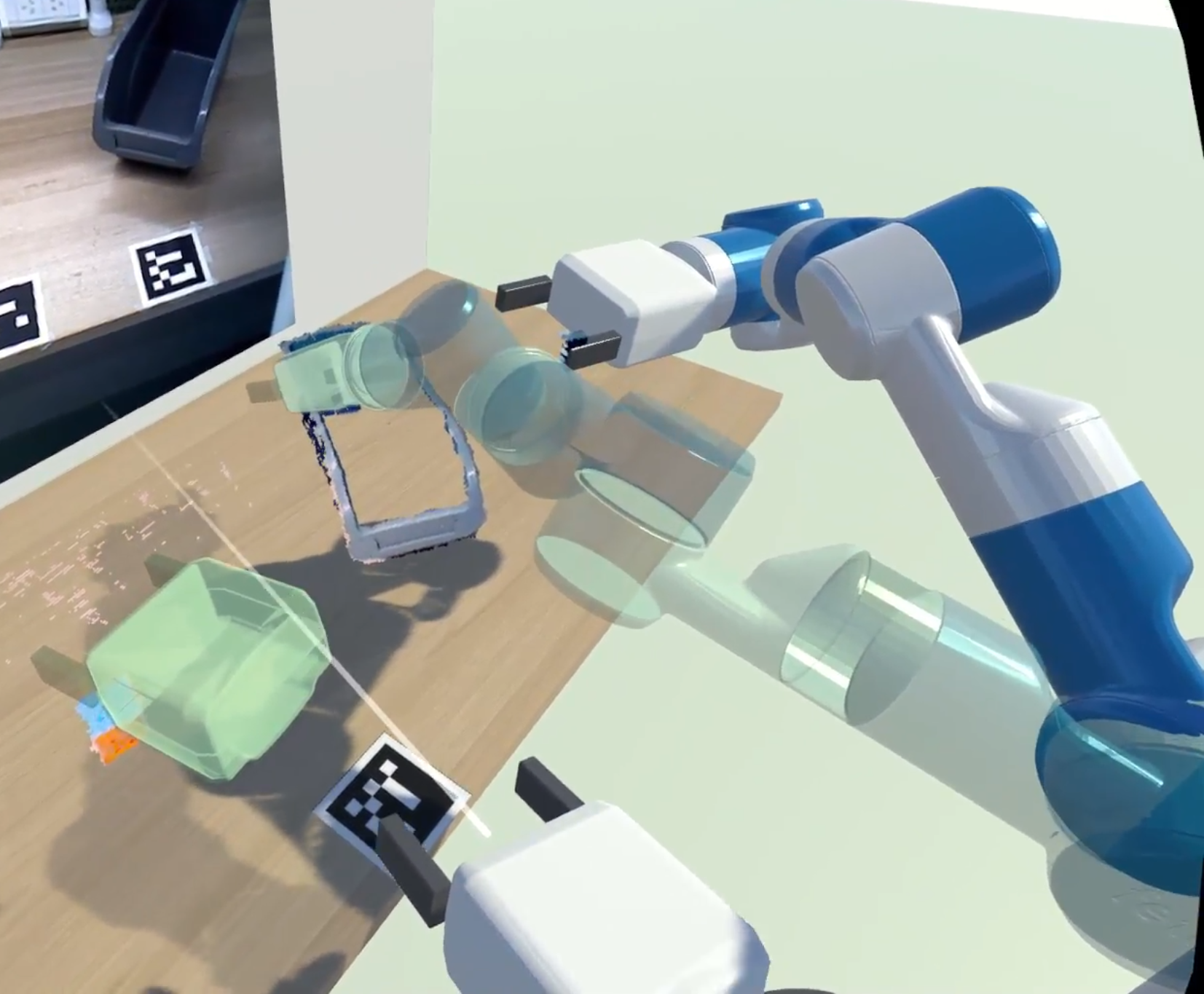}
        \caption{Operator's view}
        \label{figure:plan_b}
    \end{subfigure}
    \caption{The planning process. The semi-transparent green gripper indicates a successful plan, the trajectory of which is shown by the semi-transparent green arm in an animation loop.}
    \label{figure:plan}
\end{figure}

The plan is visualized as an animated, green, semi-transparent copy of the robot arm. When the \textit{Execute} button is pressed, if a valid motion plan has been found, then the robot attempts to execute the plan. The user is also free to not execute the plan, and instead attempt to create a new plan. A created plan is shown in Figure \ref{figure:plan}.

The \textit{Gripper} button toggles the gripper between the open and closed state. When the gripper comes in contact with an object, the force applied by the gripper is restricted by a set grip force limit.

\section{Demonstration task and discussion}

The developed system was tested in a grasping task, involving picking up an item placed on top of the table and dropping it into a box, also on the table. The object was chosen to be a Rubik's cube as it was a fitting size for the gripper, hard and would not deform when grasped, and colorful and easily visible as a point cloud. Both naive and expert VR users were invited to attempt the pick-and-place task, they were generally successful in completing the task irrespective of their prior experiences with VR. 

The trials revealed various shortcomings in the implemented program. Firstly, as pointed out previously in Sec. \ref{visualization}, the visualized point cloud was incomplete due to the occlusions as the robot camera could only capture the scene from its fixed position. This required users to make their own judgement on the complete shape based on their recognition of the item. In this work, the 2D camera footage was provided to aid the recognition. Alternative approaches may include using machine learning to classify and predict the pose of the item \cite{wonsick2021telemanipulation}, though this is limited by the amount of classes that can be identified and modelled by the program, as well as heavier computational costs. Multiple point clouds can also be combined, from different camera sources jointly localised by, for example, observing a common ArUco marker. This requires both cameras to be very accurately calibrated both in terms of their point cloud localisation and the pose detection of the ArUco marker. Since the Fetch is a mobile platform, it may also be possible to survey the scene from different vantage points with the robot's camera. This approach could be restricted by requirements for time and an allowing physical environment.

When planning motion to a goal pose using the MoveIt motion planner, the arm may stop in a position where it blocks the line of sight between the robot's camera and the item that needs to be displayed to the user. In our implementation, this required the user to be mindful when creating a trajectory for the arm, complicating the process. With assistance of the 2D image display, the user would need to plan a large movement from outside the robot's line of sight directly on to the item, or attempt to find non-obstructing poses by repeated trials. The motion planning scheme would therefore need to be improved. Potential techniques including adding a collision object representing the robot's line of sight into the MoveIt planning scene, or implementing another motion controller capable of avoiding self-occlusions \cite{he2022visibility}.

\section{Conclusion and Future work}

In this work, we present a VR interface for semi-autonomously controlling a robotic manipulator. The methods of implementation, design choices and instructions for using the system have been detailed. Currently the system is capable of teleoperation of the Fetch's arm movements with semi-autonomous motion planning, by an operator connected to the robot on a wired network. Visual information is conveyed to the operator through a mix of 2D and 3D displays, and modelled static objects.

From the system developed in this work, further efforts may be focused to address the aforementioned limitations, or to extend its application to a larger variety of robot platforms and performed tasks. Different robots may have different components and capabilities. It may be possible to devise a set of basic features required in VR teleoperation, which can be included as part of a general framework for more specific functions to be built upon, analogous to a AR visualization framework for robotics \cite{hoang2022arviz}. 

For such a purpose, said basic features would need to be implemented, notably navigation controls in the VR environment. This would also require modifications and improvements to the current manipulation system. For the robot to be mobile as well as truly remotely operated, it cannot be connected to the remote operator through a wired network, and data transported on the network would need to be scaled accordingly. This is most significant for the point cloud data, which is large in size and require frequent updates. 

A mobile robot would also allow more positions to initiate a grasping action from, which may be decided by the user or algorithmically generated. While some form of autonomy that allow the robot to control such tasks are often implemented \cite{murphy2019introduction}, there is no consensus on how much autonomy these robots should have for navigation or manipulation tasks \cite{atkeson2018happened}. To conclude, Virtual Reality offers a promising interface for semi-autonomous teleoperation of robots, however, further investigation is needed in the the division of control responsibilities between the robot and the operator, mental load of the operator compared to alternative interfaces, information visualization, safe control and experimental sensing modalities such as haptic gloves \cite{perret2018touching}.

\bibliographystyle{IEEEtran}
\bibliography{references.bib}

\begin{thebibliography}{10}
\providecommand{\url}[1]{#1}
\csname url@samestyle\endcsname
\providecommand{\newblock}{\relax}
\providecommand{\bibinfo}[2]{#2}
\providecommand{\BIBentrySTDinterwordspacing}{\spaceskip=0pt\relax}
\providecommand{\BIBentryALTinterwordstretchfactor}{4}
\providecommand{\BIBentryALTinterwordspacing}{\spaceskip=\fontdimen2\font plus
\BIBentryALTinterwordstretchfactor\fontdimen3\font minus
  \fontdimen4\font\relax}
\providecommand{\BIBforeignlanguage}[2]{{%
\expandafter\ifx\csname l@#1\endcsname\relax
\typeout{** WARNING: IEEEtran.bst: No hyphenation pattern has been}%
\typeout{** loaded for the language `#1'. Using the pattern for}%
\typeout{** the default language instead.}%
\else
\language=\csname l@#1\endcsname
\fi
#2}}
\providecommand{\BIBdecl}{\relax}
\BIBdecl

\bibitem{tanaka2018development}
M.~Tanaka, M.~Nakajima, Y.~Suzuki, and K.~Tanaka, ``Development and control of
  articulated mobile robot for climbing steep stairs,'' \emph{IEEE/ASME
  Transactions on Mechatronics}, vol.~23, no.~2, pp. 531--541, 2018.

\bibitem{klamt2018supervised}
T.~Klamt, D.~Rodriguez, M.~Schwarz, C.~Lenz, D.~Pavlichenko, D.~Droeschel, and
  S.~Behnke, ``Supervised autonomous locomotion and manipulation for disaster
  response with a centaur-like robot,'' in \emph{2018 IEEE/RSJ International
  Conference on Intelligent Robots and Systems (IROS)}.\hskip 1em plus 0.5em
  minus 0.4em\relax IEEE, 2018, pp. 1--8.

\bibitem{megalingam2018ros}
R.~K. Megalingam, D.~Nagalla, R.~K. Pasumarthi, V.~Gontu, and P.~K. Allada,
  ``Ros based, simulation and control of a wheeled robot using gamer’s
  steering wheel,'' in \emph{2018 4th International Conference on Computing
  Communication and Automation (ICCCA)}.\hskip 1em plus 0.5em minus 0.4em\relax
  IEEE, 2018, pp. 1--5.

\bibitem{hudson2021heterogeneous}
N.~Hudson, F.~Talbot, M.~Cox, J.~Williams, T.~Hines, A.~Pitt, B.~Wood,
  D.~Frousheger, K.~L. Surdo, T.~Molnar \emph{et~al.}, ``Heterogeneous ground
  and air platforms, homogeneous sensing: Team csiro data61's approach to the
  darpa subterranean challenge,'' \emph{arXiv preprint arXiv:2104.09053}, 2021.

\bibitem{atkeson2018happened}
C.~G. Atkeson, P.~B. Benzun, N.~Banerjee, D.~Berenson, C.~P. Bove, X.~Cui,
  M.~DeDonato, R.~Du, S.~Feng, P.~Franklin \emph{et~al.}, ``What happened at
  the darpa robotics challenge finals,'' in \emph{The DARPA Robotics Challenge
  Finals: Humanoid Robots to the Rescue}.\hskip 1em plus 0.5em minus
  0.4em\relax Springer, 2018, pp. 667--684.

\bibitem{stotko2019vr}
P.~Stotko, S.~Krumpen, M.~Schwarz, C.~Lenz, S.~Behnke, R.~Klein, and
  M.~Weinmann, ``A vr system for immersive teleoperation and live exploration
  with a mobile robot,'' in \emph{2019 IEEE/RSJ International Conference on
  Intelligent Robots and Systems (IROS)}.\hskip 1em plus 0.5em minus
  0.4em\relax IEEE, 2019, pp. 3630--3637.

\bibitem{whitney2020comparing}
D.~Whitney, E.~Rosen, E.~Phillips, G.~Konidaris, and S.~Tellex, ``Comparing
  robot grasping teleoperation across desktop and virtual reality with ros
  reality,'' in \emph{Robotics Research}.\hskip 1em plus 0.5em minus
  0.4em\relax Springer, 2020, pp. 335--350.

\bibitem{lemasurier2021semi}
G.~LeMasurier, J.~Allspaw, and H.~A. Yanco, ``Semi-autonomous planning and
  visualization in virtual reality,'' \emph{arXiv preprint arXiv:2104.11827},
  2021.

\bibitem{hetrick2020comparing}
R.~Hetrick, N.~Amerson, B.~Kim, E.~Rosen, E.~J. de~Visser, and E.~Phillips,
  ``Comparing virtual reality interfaces for the teleoperation of robots,'' in
  \emph{2020 Systems and Information Engineering Design Symposium
  (SIEDS)}.\hskip 1em plus 0.5em minus 0.4em\relax IEEE, 2020, pp. 1--7.

\bibitem{hoang2022arviz}
K.~C. Hoang, W.~P. Chan, S.~Lay, A.~Cosgun, and E.~A. Croft, ``Arviz: An
  augmented reality-enabled visualization platform for ros applications,''
  \emph{IEEE Robotics \& Automation Magazine}, vol.~29, no.~1, pp. 58--67,
  2022.

\bibitem{gu2021seeing}
M.~Gu, A.~Cosgun, W.~P. Chan, T.~Drummond, and E.~Croft, ``Seeing thru walls:
  Visualizing mobile robots in augmented reality,'' in \emph{2021 30th IEEE
  International Conference on Robot \& Human Interactive Communication
  (RO-MAN)}.\hskip 1em plus 0.5em minus 0.4em\relax IEEE, 2021, pp. 406--411.

\bibitem{waymouth2021demonstrating}
B.~Waymouth, A.~Cosgun, R.~Newbury, T.~Tran, W.~P. Chan, T.~Drummond, and
  E.~Croft, ``Demonstrating cloth folding to robots: Design and evaluation of a
  2d and a 3d user interface,'' in \emph{2021 30th IEEE International
  Conference on Robot \& Human Interactive Communication (RO-MAN)}.\hskip 1em
  plus 0.5em minus 0.4em\relax IEEE, 2021, pp. 155--160.

\bibitem{newbury2021visualizing}
R.~Newbury, A.~Cosgun, T.~Crowley-Davis, W.~P. Chan, T.~Drummond, and E.~Croft,
  ``Visualizing robot intent for object handovers with augmented reality,''
  \emph{arXiv preprint arXiv:2103.04055}, 2021.

\bibitem{gu2022ar}
M.~Gu, E.~Croft, and A.~Cosgun, ``Ar point\&click: An interface for setting
  robot navigation goals,'' \emph{arXiv preprint arXiv:2203.15219}, 2022.

\bibitem{coleman2014moveit}
D.~Coleman, I.~Sucan, S.~Chitta, and N.~Correll, ``Reducing the barrier to
  entry of complex robotic software: a moveit! case study,'' \emph{Journal of
  Software Engineering for Robotics}, 2014.

\bibitem{wonsick2021human}
M.~Wonsick and T.~Pad{\i}r, ``Human-humanoid robot interaction through virtual
  reality interfaces,'' in \emph{2021 IEEE Aerospace Conference (50100)}.\hskip
  1em plus 0.5em minus 0.4em\relax IEEE, 2021, pp. 1--7.

\bibitem{wonsick2021telemanipulation}
M.~Wonsick, T.~Keleștemur, S.~Alt, and T.~Pad{\i}r, ``Telemanipulation via
  virtual reality interfaces with enhanced environment models,'' in \emph{2021
  IEEE/RSJ International Conference on Intelligent Robots and Systems
  (IROS)}.\hskip 1em plus 0.5em minus 0.4em\relax IEEE, 2021, pp. 2999--3004.

\bibitem{macias2019measuring}
M.~Macia{\'s}, A.~D{\k{a}}browski, J.~Fra{\'s}, M.~Karczewski, S.~Puchalski,
  S.~Tabaka, and P.~Jaroszek, ``Measuring performance in robotic teleoperation
  tasks with virtual reality headgear,'' in \emph{Conference on
  Automation}.\hskip 1em plus 0.5em minus 0.4em\relax Springer, 2019, pp.
  408--417.

\bibitem{wei2021multi}
D.~Wei, B.~Huang, and Q.~Li, ``Multi-view merging for robot teleoperation with
  virtual reality,'' \emph{IEEE Robotics and Automation Letters}, vol.~6,
  no.~4, pp. 8537--8544, 2021.

\bibitem{quigley2009ros}
M.~Quigley, K.~Conley, B.~Gerkey, J.~Faust, T.~Foote, J.~Leibs, R.~Wheeler,
  A.~Y. Ng \emph{et~al.}, ``Ros: an open-source robot operating system,'' in
  \emph{ICRA workshop on open source software}, vol.~3, no. 3.2.\hskip 1em plus
  0.5em minus 0.4em\relax Kobe, Japan, 2009, p.~5.

\bibitem{unity2022}
\BIBentryALTinterwordspacing
U.~Technologies. (2022) Unity robotics hub. [Online]. Available:
  \url{https://github.com/Unity-Technologies/Unity-Robotics-Hub}
\BIBentrySTDinterwordspacing

\bibitem{tcp2022}
------, ``{How does the TCP Endpoint compare to Rosbridge Server?}''
  \url{https://github.com/Unity-Technologies/Unity-Robotics-Hub/blob/main/faq.md#how-does-the-tcp-endpoint-compare-to-rosbridge-server},
  2022, [Online; accessed 20-May-2022].

\bibitem{rosbridge2022}
\BIBentryALTinterwordspacing
R.~W. Tools. (2022) Ros bridge. [Online]. Available:
  \url{https://github.com/RobotWebTools/rosbridge\_suite}
\BIBentrySTDinterwordspacing

\bibitem{garrido2014automatic}
S.~Garrido-Jurado, R.~Mu{\~n}oz-Salinas, F.~J. Madrid-Cuevas, and M.~J.
  Mar{\'\i}n-Jim{\'e}nez, ``Automatic generation and detection of highly
  reliable fiducial markers under occlusion,'' \emph{Pattern Recognition},
  vol.~47, no.~6, pp. 2280--2292, 2014.

\bibitem{he2022visibility}
K.~He, R.~Newbury, T.~Tran, J.~Haviland, B.~Burgess-Limerick, D.~Kuli{\'c},
  P.~Corke, and A.~Cosgun, ``Visibility maximization controller for robotic
  manipulation,'' \emph{arXiv preprint arXiv:2202.12557}, 2022.

\bibitem{murphy2019introduction}
R.~R. Murphy, \emph{Introduction to AI robotics}.\hskip 1em plus 0.5em minus
  0.4em\relax MIT press, 2019.

\bibitem{perret2018touching}
J.~Perret and E.~Vander~Poorten, ``Touching virtual reality: a review of haptic
  gloves,'' in \emph{ACTUATOR 2018; 16th International Conference on New
  Actuators}.\hskip 1em plus 0.5em minus 0.4em\relax VDE, 2018, pp. 1--5.

\end{thebibliography}

\end{document}